\documentclass[11pt,twoside]{article}
\usepackage{clin-macros}
\usepackage{latexsym}
\usepackage{psfig}

\title{
  {\huge Completeness of Compositional Machine
         Translation for Context-Free Grammars}
  \thanks{
    The research presented here is part of my PhD-project on the completeness
    of compositional machine translation.
    In this PhD-project I address the completeness issue for several grammar
    formalisms, describing and comparing them in terms of an abstract,
    algebraic formulation of compositional grammars and compositional
    translation.
    This paper is restricted mainly to the context-free grammar formalism.
  }
}
 
\author{
  {\bf Willem-Olaf Huijsen}                                          \\
  {\em Research Institute for Language and Speech (O.T.S.)}          \\
  {\em University of Utrecht,\ \ Trans 10,\ \ NL-3512 JK,\ \ Utrecht,
                    \ \ The Netherlands}   \\
  {\em Tel. +31-30-2536057/2536178\ \ Fax. +31-30-2536000}           \\
  {\em E-mail\ \ {\tt\small Willem-Olaf.Huijsen@let.ruu.nl}}         \\
  {\em World-Wide Web\ \
       {\tt\small http://wwwots.let.ruu.nl/$\sim$Willem-Olaf.Huijsen} }
}
 
\begin{document}
 
\maketitle
 
\begin{abstract}\noindent
A machine translation system is said to be {\em complete} if all expressions
that are correct according to the source-language grammar can be translated
into the target language. This paper addresses the completeness issue for
compositional machine translation in general, and for compositional machine
translation of context-free grammars in particular.
Conditions that guarantee translation completeness of context-free grammars
are presented.
\end{abstract}

\section{Introduction}

Systems for translation of controlled language\footnote{For more information
on controlled language, see {\tt http://wwwots.let.ruu.nl/Controlled-languages/}.}
require the source text to be expressed within severe syntactical and lexical
limits. One of the objectives of such
systems is that an author who fully conforms to the imposed restrictions is
rewarded with a reliable and fully automatic translation of his text into one or
more target languages. Therefore a proof of their {\em completeness} is of
great importance. A machine translation system is said to be {\em complete}
if all expressions that are correct according to the source-language grammar
can be translated into the target language.\\
The starting point of this research has been the compositional approach to
machine translation developed in the Rosetta project, \cite{Rosetta-1994}.
An important difference is that Rosetta made use of a rather complex grammar
formalism, M-grammars, for which completeness could not be proven, whereas
the current research focuses on the provability of completeness for relatively
simple grammar formalisms, which may be more appropriate for machine
translation of controlled languages.\\
First sections~\ref{sec-compositional-grammars}
and~\ref{sec-compositional-machine-translation} describe our view and
definitions of respectively compositional grammar and compositional machine
translation.
Section~\ref{sec-completeness} presents the theme of this paper,
viz.\,{\em completeness} of compositional machine translation.
Subsequently section~\ref{sec-completeness-for-CFG} works out completeness
conditions for compositional grammars based on context-free grammars.
These conditions are rather restrictive and may therefore find application
primarily in areas such as controlled languages.
One of the objectives of ongoing research is to relax the conditions.
Section~\ref{sec-discussion} concludes the paper and discusses ongoing and
future research.

\section{Compositional Grammars}
\label{sec-compositional-grammars}%

This section defines {\em compositional grammars}
(subsection \ref{subsec-def-compositional-grammars}),
and the auxiliary notions {\em syntactic derivation tree}
(subsection \ref{subsec-def-syndertree})
and {\em semantic derivation tree} (subsection \ref{subsec-def-semdertree}).

\subsection{A Definition of Compositional Grammars}
\label{subsec-def-compositional-grammars}%

Compositional machine translation assumes that the source language (SL)
and the target language (TL) are defined by means of compositional grammars,
i.e.\,grammars that obey the well-known compositionality principle
(cf.~\cite[p.315ff]{Partee-et-al-1993,Janssen-1986,Gamut-1991}).
Abstracting away from the details of any specific syntactic formalism,
we define a {\em compositional grammar} $G$ as consisting of {\em(i)} a syntactic
component, {\em(ii)} a semantic component, and {\em(iii)} an interpretation from
the syntactic component to the semantic component (cf.\,Montague's Universal
Grammar, \cite{Thomason-1974a}).
Roughly, the syntactic component consists of a set of basic expressions (words),
each having a syntactic category, and a set of syntactic rules that build larger
expressions from basic expressions.
Likewise, the semantic component consists of a set of basic meanings,
each having a semantic category, and a set of semantic rules that build larger
meanings from basic meanings.
The interpretation associates with every basic expression a {\em set} of
basic meanings, and with every syntactic rules a {\em set} of semantic rules.\\
There now follows a more detailed description of these components, which the
eager reader may wish to skip on a first pass.\\

\bullit\
The {\em syntactic component} specifies
a finite set of basic expressions \BE{},
a finite set of syntactic rules \SynR{},
a finite set of syntactic categories \SynKats{},
and a syntactic type-assignment function \SynType{}{\cdot}.
{\em Basic expressions} are, roughly, the smallest meaningful units in a
language (more or less the stems of content words).
{\em Syntactic rules} are operations that recursively build
{\em derived expressions} from basic expressions.
{\em Syntactic categories} describe the syntactic properties of expressions.
Basic expressions~$b$ all have a syntactic category \SynCat{}{b};
syntactic rules restrict their arguments in their categories,
and specify the category of the derived expression they yield.
The {\em syntactic type-assignment function} associates every
syntactic rule~$R$ with a 2-tuple \SynType{}{R} consisting of a so-called
{\em argument list} \SynAL{}{R} of the categories of its arguments
and its resultant category.
The arity \arity{}{R} of a syntactic rule is the number of categories
in the rule's argument list.
We require that all syntactic and semantic rules are total:
They must be applicable for any combination of arguments that matches
their argument lists. Note that this is not a real restriction of
expressiveness: Any partial function can be made into a total function
by an appropriate tuning of the set of categories.

\bullit\
The {\em semantic component} has the same structure as the syntactic component:
It specifies a finite set of basic meanings \BM{},
a finite set of semantic rules \SemR{},
a finite set of semantic categories \SemKats{},
and a semantic type-assignment function \SemType{}{\cdot}.
{\em Basic meanings} are expressions of the semantic domain of some
logical language.
{\em Semantic rules} are operations in the logical language that recursively build
{\em derived meanings} from basic meanings.
For the purpose of compositional translation the choice of this logical language
is not very important. However, the semantic rules must be total.
{\em Semantic categories} describe the semantic properties of semantic expressions.
Basic meanings~$m$ all have a semantic category \SemCat{}{m};
semantic rules restrict their arguments in their semantic categories,
and specify the category of the derived meaning they yield.
The {\em semantic type-assignment function} associates every
semantic rule~$M$ with a 2-tuple \SemType{}{M} consisting of a so-called
{\em argument list} \SemAL{}{M} of the categories of its arguments
and its resultant category.
The arity \arity{}{M} of a semantic rule is the number of categories
in the rule's argument list.

\bullit\
The {\em interpretation}, denoted $\Mean{.}$, associates every basic
expression with a {\em set} of basic meanings, and every syntactic
rule with a {\em set} of semantic rules.
The arities of associated syntactic and semantic rules must match.
Note that our approach differs here from Montague grammar,
in which a basic expression (syntactic rule) is associated with
{\em exactly one} basic meaning (semantic rule).
%

\subsection{Syntactic Derivation Trees}
\label{subsec-def-syndertree}%

Derivational histories of syntactic expressions are represented
using so-called syntactic derivation trees:

\definition{Syntactic Derivation Tree}
A {\em syntactic derivation tree}~$t$ is either
a tree consisting of a single node~$b$, where~$b$ is the name of
a basic expression, or a tree of the form $R[t_1\kdots t_n]$,
where~$R$ is the name of a syntactic rule, and $t_1\kdots t_n$
is an ordered list of syntactic derivation trees.\\
We define the syntactic category of a syntactic derivation tree~$t$,
denoted {\em SynCat(t)}, to be
the resultant category of its top syntactic rule.
For convenience, we will sometimes annotate syntactic derivation trees
with their syntactic category, e.g. $t:C$.\\

Intuitively one may think of a syntactic derivation tree as
the derivational history of a syntactic expression.
However, not all syntactic derivation trees actually describe
expressions: The definition given above does not require the
syntactic rules to be applicable to their arguments.
This distinction is described by the concept of well-formedness.
 
\definition{Well-Formedness of Syntactic Derivation Trees}%
\label{def-CFG-wfness}%
A syntactic derivation tree~$t$ is {\em well-formed} if and only if
it consists of a single basic expression or otherwise if all the
syntactic rules in the tree are applicable to their arguments as
specified by tree~$t$, i.e. if and only if for all the syntactic rules
in tree $t$ {\em(i)} the number of arguments (subtrees) matches the
rule's arity, and {\em(ii)} the arguments satisfy any conditions on the
syntactic categories that may be made by the syntactic rule.\\
 
Since there is generally more than one way to derive an expression,
expressions are in general assigned a {\em set} of corresponding
syntactic derivation trees.

\subsection{Semantic Derivation Trees}
\label{subsec-def-semdertree}%

The meaning of a derived expression is derived in parallel with the syntactic
derivation process. Thus this semantic derivation process may be represented
in a tree with the same geometry as the syntactic derivation tree, but labelled
by basic meanings and semantic rules.
This tree is called a {\em semantic derivation tree}.
 
\definition{Semantic Derivation Tree}
A {\em semantic derivation tree}~$d$ is either
a tree consisting of a single node~$m$, where $m$ is the name of a basic
meaning, or a tree of the form~$M[d_1\kdots d_n]$, where~$M$ is the
name of a semantic rule, and $d_1\kdots d_n$ is an ordered list of
semantic derivation trees.\\
We define the semantic category of a semantic derivation tree~$d$,
denoted {\em SemCat(d)}, to be
the resultant category of its top semantic rule.
Semantic derivation trees may also be annotated with their semantic
category, e.g. $d:C$.\\

Since every syntactic derivation tree is associated with a {\em set}
of semantic derivation trees, every syntactic derivation tree is
associated with a {\em set} of semantic derivation trees.\\
A semantic derivation tree is well-formed if and only if there is a
corresponding well-formed syntactic derivation tree.

\section{Compositional Machine Translation}
\label{sec-compositional-machine-translation}

In our definition of compositional translation the semantic component is used
as an interlingua: Source- and target-language expressions are
{\em translation-equivalent} if and only if they have at least one well-formed
semantic derivation tree in common.

\definition{Compositional Translation}
For two compositional grammars~$G$ and~$G'$, the {\em compositional translation}
of a source-language utterance~$e$ is a set of target-language utterances,
derived as follows:

\pictura{The Process of Compositional Translation}
        {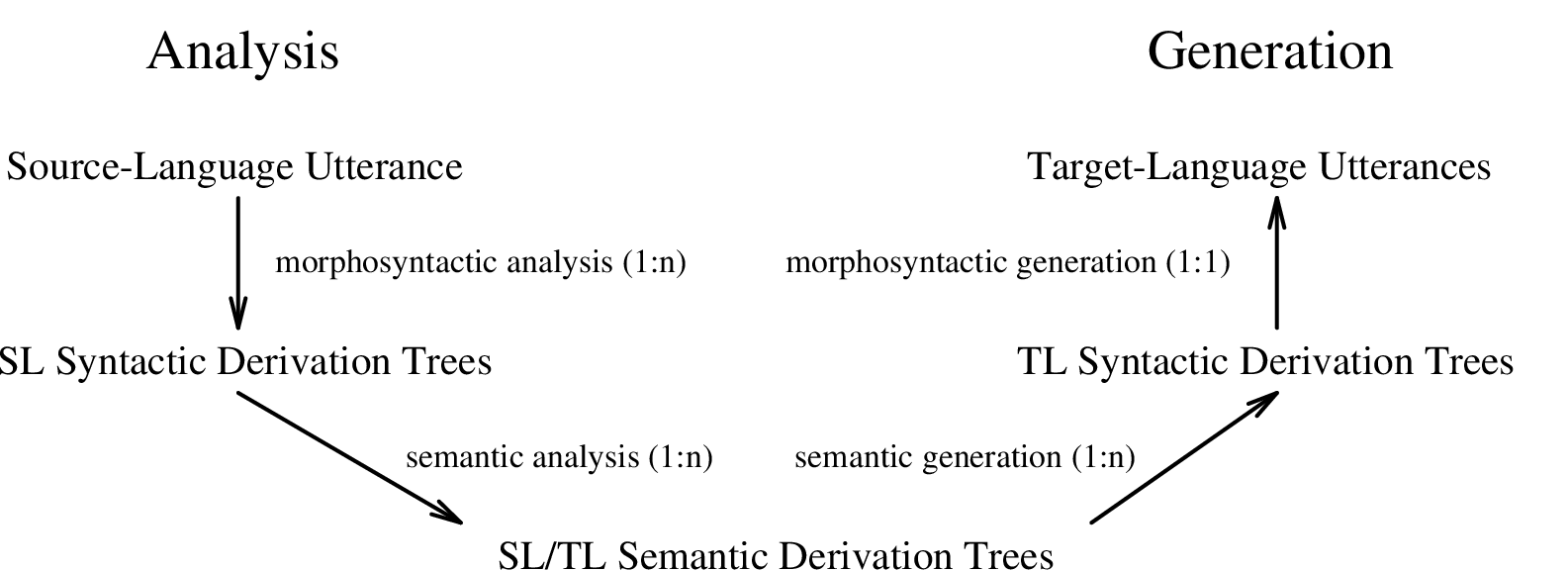,height=4cm,width=11cm}

$\bullet$ {\bf Morphosyntactic Analysis} --
  Morphosyntactic analysis performs morphological and syntactic analysis
  of a SL utterance, yielding the set of all syntactic derivation trees
  that correspond to the utterance:\\
  \mm{3} $morsynan(e)=\{b\mid b=e,\elem{b}{\BE{}}\}$\\
  \mm{3} \mm{5} $\cup\ \{R\boom{t_1\kdots t_n}\mid e=R(e_1\kdots e_n),
                  \forall i\ (1\leq i\leq n)\ \elem{t_i}{morsynan(e_i)},
                  \elem{R}{\SynR{}}\}$\\
  (`$R(e_1\kdots e_n)$' denotes the result of applying rule $R$
   to expressions $e_1\kdots e_n$).\\
$\bullet$ {\bf Semantic Analysis} --
  Semantic analysis of a syntactic derivation tree yields the set of all
  corresponding semantic derivation trees:\\
  \mm{3} $seman(b)=\Mean{b}$\\
  \mm{3} $seman(R\boom{t_1\kdots t_n})=\{M\boom{d_1\kdots d_n}\mid
          \elem{M}{\Mean{R}}\ \wedge\ 
          \forall i\ (1\leq i\leq n)\ \elem{d_i}{seman(t_i)}\}$\\
$\bullet$ {\bf Semantic Generation} --
  Semantic generation from a semantic derivation tree yields the set of all
  corresponding syntactic derivation trees:\\
  \mm{3} $semgen(m)=\{b\mid\elem{m}{\Mean{b}}\}$\\
  \mm{3} $semgen(M\boom{d_1\kdots d_n})=\{R\boom{t_1\kdots t_n}\mid
          \elem{M}{\Mean{R}}\ \wedge\ 
          \forall i\ (1\leq i\leq n)\ \elem{t_i}{semgen(d_i)}\}$\\
$\bullet$ {\bf Morphosyntactic Generation} --
  Morphosyntactic generation for a {\em well-formed} syntactic derivation tree
  produces the corresponding utterance:\\
  \mm{3} $morsyngen(b)=b$\\
  \mm{3} $morsyngen(R\boom{t_1\kdots t_n})=R(e_1\kdots e_n)$,
         where $\forall i\ (1\leq i\leq n)\ \elem{e_i}{morsyngen(t_i)}$

\section{Completeness of Machine Translation}
\label{sec-completeness}
 
An important question regarding the reliability of compositional translation is
what we call the {\em completeness}\,\footnote{The term `completeness' was taken
from \cite[pp.342-343]{Whitelock-1994}. In the Rosetta framework completeness is
known as `strict isomorphism', and is discussed in~\cite{Landsbergen-1987a}
and~\cite{Rosetta-1994}.} issue:
{\em Can the translation process be guaranteed to produce at least one translation?}
In subsection~\ref{sss-three-levels} we first make this notion of
completeness precise.
Then, in subsection~\ref{sss-guaranteeing-completeness}, we investigate
what conditions must be satisfied to guarantee completeness.
In section~\ref{sec-completeness-for-CFG} conditions are elaborated for
compositional grammars based on context-free grammars.
 
\subsection{Three Levels of Completeness}\label{sss-three-levels}%
Completeness is about the guaranteed generation of well-formed translations,
given a specific SL and TL grammar, and translation process.
However, this description does not make precise from which stage on the
translation process must be guaranteed to succeed.
Depending on this, one may distinguish (at least) three levels of completeness
(cf.\,fig.\,1):\\
\bullit\ {\bf Utterance Completeness} --
  For each well-formed SL utterance, the translation process yields at least one
  well-formed TL utterance.\\
\bullit\ {\bf Syntactic Completeness} --
  For each syntactic derivation tree of each well-formed SL utterance, the
  translation process yields at least one well-formed TL utterance.\\
\bullit\ {\bf Semantic Completeness} --
  For each semantic derivation tree of each syntactic derivation tree
  of each well-formed SL utterance, the translation process yields at
  least one well-formed TL utterance.\\
{\bf Note: } Semantic completeness subsumes syntactic completeness, which
             in turn subsumes utterance completeness.\\
 
Naively, one would like a machine translation system to produce at least one
translation for every SL utterance. This requirement is included in the
definition of utterance completeness above.
However, it is well-known that natural-language utterances are often
ambiguous. For each of its interpretations, such an ambiguous utterance may
have a different translation. Therefore, a machine translation system should
be able to provide at least one translation {\em for each of the
interpretations} of the SL utterance.
Natural-language ambiguity takes on two forms: structural ambiguity and lexical
ambiguity. The notion of syntactic completeness takes
care of the structural ambiguity: It is formulated in terms of structurally
unambiguous syntactic derivation trees.
However, syntactic completeness is still unsatisfactory,
as syntactic derivation trees are often lexically ambiguous. This is
due to the fact that basic expressions may have more than one meaning,
and syntactic rules may have more than one semantic rule associated with them.
What is needed is a formulation of completeness in terms of a structure that
is both structurally and lexically unambiguous. The solution is provided by
the notion of semantic completeness.
Therefore, from now on, the term `completeness' will be taken to refer to
semantic completeness only.
 
\definition{Completeness}
For a pair of compositional grammars \tuple{G,G'},
compositional translation from~$G$ to~$G'$ is {\em complete} if and only if
for each well-formed semantic derivation tree,
the translation process yields at least one well-formed TL utterance.

\subsection{Guaranteeing Completeness}\label{sss-guaranteeing-completeness}%
 
The central issue of this paper is the question of how to guarantee completeness.
Or stated in terms of the process of compositional translation described above:
What conditions on the SL and TL grammars are sufficient (and necessary) to
guarantee that, after successful analysis, generation can produce a well-formed
TL expression?
Generation comprises semantic generation and morphosyntactic generation
(cf.\,fig.\,1).\\

{\bf Completeness of Morphosyntactic Generation} --
Morphosyntactic generation evaluates the syntactic derivation trees yielded
by semantic generation by recursive rule application.
As stated in section~\ref{sec-compositional-grammars}, we assume that all
syntactic rules are total for the categories of their arguments.
Rule application therefore succeeds if and only if the arguments are of
the correct categories. To ensure this we must move upstream to
semantic generation.\\
 
{\bf Completeness of Semantic Generation} --
Semantic generation simply replaces the basic meanings and semantic rules in
the semantic derivation tree with corresponding syntactic elements
of the TL grammar, forming the TL syntactic derivation trees.
An obvious necessary and sufficient condition for completeness of semantic
generation is that there be
{\em at least one} translation-equivalent counterpart in the TL grammar
for each possible semantic element in the SL semantic derivation trees.
A compositional grammar pair satisfying this condition is called a
{\em homomorphic grammar pair} (see also~\cite[p.368]{Rosetta-1994}):

\definition{Grammar Homomorphism}%
\label{def-grammar-homomorphism2}%
A compositional grammar pair \tuple{G,G'} is {\em homomorphic from~$G$ to~$G'$}
if and only if~$G'$ is {\em attuned} to~$G$:
\begin{romanlist}
\item For each SL basic expression~$b$,
      for each of the basic meanings~$m$ of~$b$,
      there is at least one TL basic expression~$b'$ such that
      basic meaning~$m$ is also a basic meaning of~$b'$.
      Formally,
      $\forall\elem{b}{\BE{}}\ \forall\elem{m}{\Mean{b}}
       \ \exists\elem{b'}{\BE{}}\ \ \elem{m}{\Mean{b'}}$.
\item For each SL syntactic rule~$R$,
      for each of the semantic rules~$M$ of~$R$,
      there is at least one TL syntactic rule~$R'$ such that
      semantic rule~$M$ is also a semantic rule of~$R'$.
      Formally,
      $\forall\elem{R}{\SynR{}}\ \forall\elem{M}{\Mean{R}}
       \ \exists\elem{R'}{\SynR{}}\ \elem{M}{\Mean{R'}}$.
\end{romanlist}

However, to demand grammar homomorphism is only a necessary condition
for completeness, and not a sufficient one.
It merely guarantees that for every well-formed SL semantic derivation
tree there is a corresponding TL syntactic derivation tree, and does
not guarantee that this syntactic derivation tree is well-formed.
The next section is about such sufficient conditions for context-free
grammars.
 
\section{Completeness for CFG-Based Compositional Grammars}
\label{sec-completeness-for-CFG}

This section presents completeness conditions for translation between
compositional grammars based on the context-free grammar (CFG) formalism.
We assume that the reader is familiar with this formalism.
Subsection~\ref{ss-cfg-based-compositional-grammar} explicates how a
compositional grammar can be based on context-free grammars.
Subsections~\ref{ss-m2o-catcor} and~\ref{ss-m2m-catcor} subsequently
develop completeness conditions for such compositional grammars.

\subsection{CFG-Based Compositional Grammar}
\label{ss-cfg-based-compositional-grammar}

A compositional grammar consists of a syntactic component with basic
expressions and syntactic rules, a semantic component with basic
meanings and semantic rules, and an interpretation from the syntactic
component to the semantic component.
Here we model the syntactic component as a CFG. The semantic component
and the interpretation are as defined above.

In the syntactic component we let basic expressions correspond to rewrite rules
that do not have right-hand side (RHS) nonterminals. The rule's RHS corresponds
to the lexical material of the basic expression; the rule's left-hand side
(LHS) nonterminal corresponds to the syntactic category of the basic expression.
We let syntactic rules correspond to rewrite rules that {\em do} have RHS
nonterminals. The type of a syntactic rule is a 2-tuple consisting of a list of
categories of the arguments it expects and the category of the expression it
produces. The list of categories corresponds to an ordered list of the rewrite
rule's RHS
nonterminals; the resultant category corresponds to the rewrite rule's LHS
nonterminal. The operation performed by the syntactic rule is the in-order
concatenation of its RHS terminals and nonterminals, where the nonterminals are
replaced with the lexical material of the expressions which are provided as
arguments to the syntactic rule. An example illustrates this:\\

{\bf Example}\ \ {\em CFG-Based Compositional Grammars}\\
In this example we briefly illustrate CFG-based compositional grammars.
Consider the following table, which shows the syntactic component of a
CFG-based compositional grammar and its interpretation in the semantic
component.
\begin{quote}
\begin{tabular}{llll}
  {\em CFG Rewrite Rule}
                   & {\em Syntactic Rule} & {\em Basic Expression}
                                                 & {\em Interpretation}\\
                   & {\em Name\,:\,Type} & {\em Name\,:\,Category} \\ \hline
  $A \raS B\ C$    & $R_1:\tuple{\tuple{B,C},A}$ & & $\{M_1\}$ \\
  $A \raS a\ B\ d$ & $R_2:\tuple{\tuple{B},A}$   & & $\{M_{2a},M_{2b}\}$ \\
  $A \raS e\ C\ B$ & $R_3:\tuple{\tuple{B,C},A}$ & & $\{M_{3a},M_{3b}\}$ \\
  $B \raS b$       & & $b:B$                       & $\{m_1\}$           \\
  $C \raS c$       & & $c:C$                       & $\{m_{2a},m_{2b}\}$
\end{tabular}
\end{quote}
Observe that the order of syntactic categories in the argument list
need not be the same as the order in the rewrite rules (see $R_1,R_3)$.
Syntactic rules $R_1$ and $R_3$ have two arguments. As a consequence
semantic rules $M_1$, $M_{3a}$ and $M_{3b}$ are binary operators.
Syntactic rule $R_2$ and semantic rules $M_{2a}$ and $M_{2b}$ have
one argument.

The notion of well-formedness can be made more precise now:

\definition{CFG-well-formedness}\label{CFG-well-formedness}%
A CFG syntactic derivation tree $t$ is {\em CFG-well-formed} if and only if
it is either the name of a basic expression, or
a tree of the form $R\boom{t_1\kdots t_n}$,
such that {\em(i)} rule $R$'s argument list matches the list of syntactic
categories of the subtrees $t_1\kdots t_n$:
$\SynAL{}{R}=\tuple{\SynCat{}{t_1}\kdots \SynCat{}{t_n}}$,
and {\em(ii)} subtrees $t_1\kdots t_n$ are CFG-well-formed.\\

What about the `translation power' of CFG-based compositional grammars?
The compositional translation method described in
section~\ref{sec-compositional-machine-translation} demands that basic
expressions
of the source language correspond to basic expressions in the target
language, and that the syntactic rules of the source-language
correspond to syntactic rules of the target language with the same arity.
This restricts the translation power considerably. The main degrees of
freedom in the translation relation are the following.
In the syntactic rules, the nonterminals need not occur in the same
order as in the argument list. This allows translation-equivalent
rules to describe word-order differences between languages.
Syntactic rules may also introduce lexical material other than that
of the arguments. This is called {\em syncategorematic introduction}
(cf. syntactic rules $R_2$ and $R_3$ in the example above, where basic
expressions $a$, $d$ and $e$ are left out).
The third degree of freedom relates to the correspondence between categories
of source- and target-language grammars.\\
Subsection~\ref{ss-m2o-catcor} now develops a completeness condition for
CFG-based compositional grammars. 
Subsection~\ref{ss-m2m-catcor} then shows that this condition is rather
restrictive and presents a way to relax it.

\subsection{CFG Completeness for Many-to-One Category Correspondence}
\label{ss-m2o-catcor}

In this section we show how a restriction of the correspondence between
syntactic and semantic categories of the target language can lead to
completeness.
First we formally define a restriction of this correspondence.

\definition{N-1 Category Correspondence}
There is an {\em N-1 category correspondence} between a semantic component
and a syntactic component of a compositional grammar if and only if
there is a function $f:\SemKats{}\raS \SynKats{}$ such that:\\
\mm{3}\bullit\ $\forall\elem{m}{\BM{}}\ \ \forall\elem{b}{\BE{}}\ \
                \elem{m}{\Mean{b}}\ \implies\ \SynCat{}{b}=f(\SemCat{}{m})$\\
\mm{3}\bullit\ $\forall\elem{M}{\SemR{}}\ \ \forall\elem{R}{\SynR{}}$\\
\mm{5} $\left(\elem{M}{\Mean{R}} \wedge \SemType{}{M}=\tuple{\tuple{c_1\kdots c_n},c}\right)\ 
        \implies\ \SynType{}{R}=\tuple{\tuple{f(c_1)\kdots f(c_n)},f(c)}$\\

The restriction of compositional grammars to such an N-1 category
correspondence together with the grammar homomorphism condition
gives us completeness:

\theorem{CFG Completeness for Many-to-One Category Correspondence}
\label{th-cfg-completeness-1}%
For any CFG-based compositional grammar pair~\tuple{G,G'}, compositional
translation from~$G$ to~$G'$ is {\em complete} if
{\em(i)} the grammar pair is homomorphic from~$G$ to~$G'$, and
{\em(ii)} there is an N-1 category correspondence between the semantic
and the syntactic categories of $G'$.\\

\begin{proof}
As we are concerned with semantic completeness,
we have to prove that for every grammatical SL~utterance, for every one of
its well-formed semantic derivation trees, there exists at least
one grammatical TL utterance.
As we assume it to be trivial that morphosyntactic generation succeeds
for CFG-well-formed syntactic derivation trees, we focus on semantic generation.
We must show that every well-formed semantic derivation tree always
yields at least one {\em CFG-well-formed} TL syntactic derivation tree.
We do this by induction on the depth of the semantic derivation trees.
 
\inductionbase
A semantic derivation tree of depth~1 is a basic meaning.
Homomorphism from~$G$ to~$G'$ guarantees that there is at least one
TL basic expression that is associated with that basic meaning.
Basic expressions are trivially CFG-well-formed syntactic derivation trees.
 
\inductionhypothesis
For every well-formed semantic derivation tree derivable in~$G$
which is of depth $m$ or less, compositional translation yields at least one
CFG-well-formed TL syntactic derivation tree in $G'$.
 
\inductionstep
Assuming the induction hypothesis holds for arbitrary depth~$m$, we must prove
that it also holds for depth~$m+1$.
Every well-formed semantic derivation tree of depth~$m+1$ is of the form
$M[d_1\kdots d_n]:A$, where each subtree~$d_i$ is of the form $M_i[\dots]:A_i$
(see fig.\,2 below).
Because of the given well-formedness of the semantic derivation tree we
know that~$M$ is applicable to its arguments, so that its argument list
\tuple{A_1\kdots A_n} matches the semantic categories of the arguments $A_i$.
Homomorphism guarantees that~$M$ has at least one associated syntactic
rule~$R'$, which has some argument list \tuple{B_1\kdots B_n}.
The induction hypothesis guarantees that every tree~$d_i$ has at least one
CFG-well-formed TL syntactic derivation tree $t_i'=R_i'[\dots]:C_i$ associated
with it. Note that the induction hypothesis says nothing about the categories
$C_i$ of these trees.
 
\pictura{Induction Step: Generating Syntactic from Semantic Derivation Trees}
        {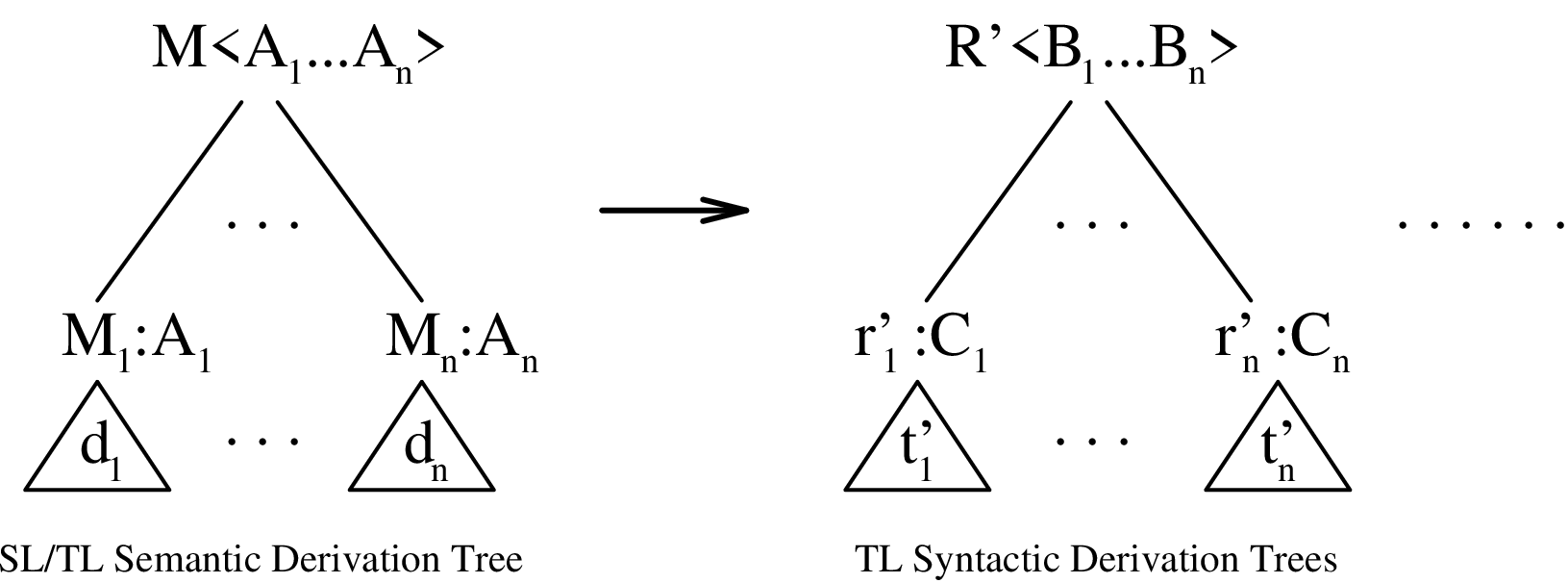,height=4cm,width=12cm}
 
The remaining question is whether there is at least one TL~syntactic derivation
tree formed in this way which is CFG-well-formed, i.e. for which
(def.\,\ref{def-CFG-wfness}):
{\em(i)} rule~$R'$ is applicable to its arguments, and
{\em(ii)} all subtrees~$t_i'$ are CFG-well-formed.
Condition {\em(ii)} is covered by the induction hypothesis.
Condition {\em(i)} requires that
the argument list of rule~$R'$ matches the syntactic categories of the
subtrees $t'_1\kdots t'_n$:
$\SynAL{}{R'}=\tuple{B_1\kdots B_n}=\tuple{\SynCat{}{t'_1}\kdots \SynCat{}{t'_n}}$.
From the condition in the theorem we know that there is an N-1 category
correspondence~$f$ between the semantic categories and the syntactic categories
of~$G'$.
Because rule~$R'$ is associated with rule~$M$, we know that for all
$1\leq i\leq n$, $B_i=f(A_i)$. Since for all $1\leq i\leq n$, we also know
that tree~$t_i'$ is associated with tree~$d_i$, it holds that $C_i=f(A_i)$.
Since $f$ is a function, it must hold that for all $1\leq i\leq n$,
$B_i=C_i$, so that the argument list of~$R'$ matches the categories
of its arguments.
Therefore, every such rule $R'$ is applicable to its arguments,
so that completeness is guaranteed.
\backupline
\end{proof}

\subsection{Many-to-Many Category Correspondence}
\label{ss-m2m-catcor}

The N-1 category correspondence condition is rather restrictive. It implies
that a semantic category of the source language must be translated into
exactly one syntactic catgory of the target language. We would
like to have a looser category correspondence. For example, consider the
following grammar rules for translating between English and French noun
phrases, where French uses agreement on determiners and nouns:
\begin{center}
  \begin{tabular}{ccc}
  {\em English Syntax} & {\em Semantics} & {\em French Syntax} \\ \hline
  $R_1 : NP\ \ra\ DET\ N$
    & $M_1 : \ul{NP}\ \ra\ \ul{DET}\ \ul{N}$
    & $R'_{1a} : NP'\ \ra\ DET'_m\ N'_m$ \\
  & & $R'_{1b} : NP'\ \ra\ DET'_f\ N'_f$
  \end{tabular}
\end{center}
Here we would like to relate semantic category $\ul{DET}$ to syntactic
categories $DET'_m$ and $DET'_f$, and semantic category $\ul{N}$ to
syntactic categories $N'_m$ and $N'_f$.
To be able to do so we could allow every semantic category to be associated
with a number of syntactic categories instead of just one.
This corresponds to an N-N category correspondence.

\definition{N-N Category Correspondence}
There is an {\em N-N category correspondence} between a semantic component
and a syntactic component of a compositional grammar if and only if
there is a function $f:\SemKats{}\raS \SynKats{}$ such that:\\
\mm{3}\bullit\ $\forall\elem{m}{\BM{}}\ \ \forall\elem{b}{\BE{}}\ \
                \elem{m}{\Mean{b}}\ \implies\ \elem{\SynCat{}{b}}{f(\SemCat{}{m})}$\\
\mm{3}\bullit\ $\forall\elem{M}{\SemR{}}\ \ \forall\elem{R}{\SynR{}}$\\
\mm{10}        $(\elem{M}{\Mean{R}}
                \wedge \SemType{}{M}=\tuple{\tuple{c_1\kdots c_n},c})\ 
                 \implies\ \SynType{}{R}=\tuple{\tuple{c_1'\kdots c_n'},c'}$,\\
\mm{10}        where $\forall i\ (1\leq i\leq n)\ \elem{c_i'}{f(c_i)}$
               and \elem{c'}{f(c)}\\
For a semantic category $C$ the set of corresponding syntactic categories
$f(C)$ is called the {\em category correspondence set} of $C$ and is
denoted \~C.\\

For this new situation we must adjust the completeness condition.
Referring to fig.\,2, it now is the case that each syntactic category $C_i$
may be any category in the set $f(A_i)$.
As the induction hypothesis guarantees only one successful translation for
each subtree~$d_i$ -- and it is not known which one -- to guarantee
completeness is to guarantee that there is a syntactic rule $R'$ for
{\em every} argument list in $f(A_1)\times\dots\times f(A_n)$.
This is an unrealistic condition: In the English/French example it
corresponds to the demand that there must be a French syntactic rule for
all four argument lists
$\tuple{DET_m,N_m},\tuple{DET_m,N_f},\tuple{DET_f,N_m},\tuple{DET_f,N_f}$.
But to demand that for example there is a syntactic rule $R'$ that combines
a masculine determiner $DET_m$ and a feminine noun $N_f$, as this would imply,
is nonsensical.
The underlying problem is that the agreement dependencies cannot be expressed
explicitly in the CFG grammar formalism.
The lesson to be learned from this example is that the
dependencies between the categories should be taken into account.
We present a way of encoding information about the dependencies
between categories in CFG-based compositional grammar.
To this end we distinguish two kinds of category correspondence.

\definition{Conjunctive/Disjunctive Correspondence Category}
For a compositional grammar, a semantic category $N$ is a {\em conjunctive
(correspondence) category} if and only if for every well-formed semantic
derivation tree $d$ of category $N$, for {\em every} corresponding category
$N'$ in {\em \~N}, there exists at least one
corresponding well-formed syntactic derivation tree $t'$ of category $N'$.
Any semantic category that is not a conjunctive correspondence category is
called a {\em disjunctive (correspondence) category}.
Semantic categories that have only one syntactic category in their category
correspondence set are trivially conjunctive categories.\\
 
For example, in the case of the English/French NP rules,
the semantic category $\ul{DET}$ corresponds {\em conjunctively} to categories
{\em DET\,$'_m$} and {\em DET\,$'_f$} (any determiner has both a masculine and
a feminine form), whilst semantic category $\ul{N}$ corresponds
{\em disjunctively} to categories {\em N\,$'_m$} and {\em N\,$'_f$}
(nouns usually have either masculine or feminine gender).
Semantic category $\ul{NP}$ corresponds to only one category, {\em NP}\,$'$,
and is therefore a conjunctive category.\\
 
How can we use this to establish a condition for completeness?
The key idea is that some of the CFG-well-formed syntactic derivation
trees of some category~$A$ may be guaranteed to translate into at least one
CFG-well-formed TL syntactic derivation tree {\em for all categories}
in~$\tilde{A}$, instead of `for at least one'.
Category~$A$ is then said to {\em correspond conjunctively} to the
categories in~$\tilde{A}$.
As opposed to disjunctive categories, a conjunctive category does not require
every rule~$R'$ to have translation-equivalent variants for all categories
in~$\tilde{A}$.
Thus, the distinction between conjunctive and disjunctive categories
allows for a more realistic condition on the grammars.\\
We adjust the definition of N-N category correspondence, taking into
account the distinction between conjunctive and disjunctive categories.\\
As for the basic meanings and basic expressions:
For every basic meaning $m$, if its category $C$ is a disjunctive
category, there must be at least one associated basic expression $b'$ with
category $C'$ {\em for at least one category} $C'$ in $\tilde{C}$.
If category $C$ of basic meaning $m$ is a conjunctive category, then there
must exist at least one associated basic expression $b'$ with category $C'$
{\em for every category} $C'$ in $\tilde{C}$.\\
As for the semantic and syntactic rules, for every semantic rule $M$
with type \tuple{\tuple{A_1\kdots A_n},A}, we establish conditions on
the syntactic rules with which they are associated.
Again referring to fig.\,2, when generating a syntactic derivation tree
from a semantic derivation tree, for subtrees $d_i$ that have a conjunctive
category $C$ we can guarantee a tree $t_i'$ {\em for every} category in
$\tilde{C}$.
For subtrees $d_i$ that have a disjunctive category $C$ we can guarantee
a tree $t_i$ {\em for only one} category in $\tilde{C}$, and we do not
know which one.
Therefore, we must guarantee that for every tuple\footnote{Consider the
following auxiliary definitions.
For any argument list \tuple{A_1\kdots A_n}, define sets $I_c$ and $I_d$
as consisting of the indices of its conjunctive and disjunctive categories,
respectively.
Define \tuple{A_i\mid\elem{i}{I_c}} and \tuple{A_i\mid\elem{i}{I_d}}
as the corresponding subtuples.}
\elem{D}{{\sf X}_{\elem{i}{I_d}}\ \tilde{A_i}} of the syntactic categories
corresponding to disjunctive categories of $M$,
there exists {\em at least one} syntactic rule $R'$ with type
\tuple{\tuple{B_1\kdots B_n},B} such that:
\begin{bulletlist}
\item The tuple of the syntactic categories corresponding to the disjunctive
      categories of the argument list of $M$ is equal to $D$:
      \tuple{B_i\mid\elem{i}{I_d}}=$D$.
\item Every syntactic category $B_i$ that corresponds to a conjunctive category
      $A_i$ of the argument list of $M$ is in the category correspondence set
      of $A_i$:
      $\forall \elem{i}{I_c}\ \ \elem{B_i}{\tilde{A_i}}$.
\item In addition, the resultant category $A$ of semantic rule $M$ must be taken
      into account. If this is a disjunctive category, then it suffices if the
      resultant category $B$ of the syntactic rule $R'$ is in $\tilde{A}$.
      If category $A$ is a conjunctive category, then there must be at least
      one syntactic rule $R'$ with resultant category $N$ for all categories
      $N$ in $\tilde{A}$.
\end{bulletlist}

Using this condition we again obtain completeness:
 
\theorem{CFG Completeness for Many-to-Many Category Correspondence}
For any CFG-based compositional grammar pair \tuple{G,G'}, compositional
translation from~$G$ to~$G'$ is {\em complete} if
{\em(i)} the grammar pair is homomorphic from~$G$ to~$G'$, and
{\em(ii)} there is an N-N category correspondence between the semantic
and the syntactic categories of $G'$, where every semantic category of $G'$
has been declared conjunctive or disjunctive and the sets of categories
of $G'$ satisfy the condition described above.\\

Because of space limitations we do not include the proof; we trust that the
description of the condition above gives the reader an insight into
how the proof can be given.\\

{\bf Example}\ \ Returning to the English/French example discussed earlier,
we declared $\ul{DET}$ a conjunctive, $\ul{N}$ a disjunctive, and $\ul{NP}$
a conjunctive category. Checking the condition formulated above, this
amounts to the requirement that for every tuple $D$ in
$\{\tuple{N_m'},\tuple{N_f'}\}$, there exists a syntactic rule $R'$ such that
\tuple{B_i\mid\elem{i}{I_d}}=$D$ and
$\forall \elem{i}{I_c}\ \ \elem{B_i}{\tilde{A_i}}$, which is indeed the case.

\section{Conclusion and Future Research}%
\label{sec-discussion}

In this paper we presented the issue of completeness for compositional translation,
and discussed how conditions for compositional translation could be found.
In section~\ref{sec-completeness-for-CFG} we examined the completeness issue for
context-free grammars. We established completeness conditions for grammars with an
N-1 category correspondence. As this condition is rather restrictive, we relaxed
this condition to an N-N category correspondence condition. The first attempt
however led to unrealistic conditions on the grammar rules, so that it was
necessary to introduce the distinction between conjunctive and disjunctive
categories. We adjusted the N-N category correspondence condition accordingly,
and obtained a completeness condition for grammars with an N-N category
correspondence.\\

The central issues in ongoing and future research are
{\em(i)}    the completeness issue for some other grammar formalisms,
{\em(ii)}   the algebraic formulation of completeness, and
{\em(iii)}  polynomial compositional translation.\\

{\bf (i)\ Completeness for Other Grammar Formalisms} --
The definite-clause grammar formalism (DCG, see e.g.\,\cite{Pereira-Shieber-1987})
extends the CFG grammar formalism with attributes added to the nonterminals.
Attributes have a variety of uses, one of the most prominent being the
enforcement of agreement relations.
As for the completeness condition for DCG, we assume the same conditions on the
nonterminals as we did for CFG. In addition, we formulate restrictions on the
use of attributes.
A proof has been established for completeness of grammars that satisfy these
restrictions.\\
%
%
Future research will also address the completeness issue for Tree-Adjoining Grammars.
Tree-Adjoining Grammars are interesting because they are somewhat more
expressive than CFG's (they are so-called mildly context-sensitive),
and it enables expressing linguistic phenomena such as long-distance
dependencies.\\

{\bf (ii)\ Algebraic Formulation of Compositional Translation} --
Compositional grammar, compositional translation and the completeness issue
can be formulated algebraically.
Such an algebraic formulation has a number of advantages:
{\em(i)} it abstracts away from the details of specific grammar formalisms,
thus revealing the essentials of compositional translation and completeness,
{\em(ii)} this abstraction provides a basis for the comparison of different
grammar formalisms, and 
{\em(iii)} an algebraic formulation gives access to well-investigated
mathematical theory, the results of which may be readily carried over.
I hope to use the algebraic formulation as a basis for the investigation
of the combination of the use of features and completeness.
For other work on algebraic description of natural language, see
\cite{Janssen-1986,Hendriks-1993}. An algebraic view on compositional
translation is presented in \cite[Ch.19]{Rosetta-1994}.\\

{\bf (iii)\ Polynomial Compositional Translation} --
Another line of work is concerned with an extension of the method of compositional
translation for grammar formalisms that use only concatenative operations.
The basic idea here is a generalization of the unit of translation-equivalence
from single elements to combinations of these (polynomials).
This improves `translation power', as it becomes possible to overcome
all kinds of translation problems due to structural divergencies between languages.
For example it becomes possible to relate a structure like $[A\ [B\ C]]$
with a structure like $[A'\ B'\ C']$.
I hope to show that, as polynomially derived algebras are algebras again,
completeness conditions found for compositional translation will carry over
to polynomial compositional translation.

 
\begin{footnotesize}

\end{footnotesize}
 
 
\end{document}